# LEGAL DOCUMENTS DRAFTING WITH FINE-TUNED PRE-TRAINED LARGE LANGUAGE MODEL

Chun-Hsien Lin and Pu-Jen Cheng

Department of Computer Science & Information Engineering, National Taiwan University, Taipei, Taiwan

*ABSTRACT*

*With the development of large-scale Language Models (LLM), fine-tuning pre-trained LLM has become a mainstream paradigm for solving downstream tasks of natural language processing. However, training a language model in the legal field requires a large number of legal documents so that the language model can learn legal terminology and the particularity of the format of legal documents. The typical NLP approaches usually rely on many manually annotated data sets for training. However, in the legal field application, it is difficult to obtain a large number of manually annotated data sets, which restricts the typical method applied to the task of drafting legal documents. The experimental results of this paper show that not only can we leverage a large number of annotation-free legal documents without Chinese word segmentation to fine-tune a large-scale language model, but more importantly, it can fine-tune a pre-trained LLM on the local computer to achieve the generating legal document drafts task, and at the same time achieve the protection of information privacy and to improve information security issues.*

## 1. INTRODUCTION

In recent years, researchers have applied neural networks to natural language processing, achieving state-of-the-art performance in processing legal documents, such as tasks related to textual entailment [1] and legal question answering [2]. With the development of large-scale Language Models (LLM), fine-tuning pretrained LLM to address natural language processing tasks, as mentioned above, has become a mainstream paradigm [3]. However, challenges persist when employing natural language text generation techniques as a solution for highly specialized tasks like legal document drafting.

One of the primary reasons for these challenges stems from the need for more training datasets for pretrained large language models to incorporate legal professional documents comprehensively. Legal statutes are subject to amendments based on different temporal and spatial contexts, affecting the legal effects described in subsequent legal documents. Moreover, training language models for the legal domain necessitates abundant legal documents to enable the model to grasp legal terminology and the specific format of legal documents. This reliance typically involves extensive human annotation, presenting practical difficulties in obtaining labeled datasets for legal applications, thus restricting the applicability of traditional NLP methods in legal document drafting tasks.

In the context of the Chinese language, prior approaches to handling Chinese data often required word segmentation pre-processing [4]. The quality of word segmentation results can significantly impact the outcomes of downstream Chinese NLP tasks. If there is no need for Chinese word segmentation, this will substantially lower the threshold for applying NLP approaches to Chinese, especially when processing legal documents. Additionally, legal documents often contain sensitive information, and legal professionals are unwilling to share documents they draft with third-party information systems. This reluctance is due to uncertainties regarding how the information sent out is processed, where it is stored, and whether the data is stored in the database of the artificial intelligence system. Thus, when developing relevant application systems for the legal profession, considerations must include information privacy and security concerns.

To address these challenges, this paper presents a process for fine-tuning a large-scale language model, enabling users to perform fine-tuning on a large pretrained language model locally and apply it to legal document drafting. The proposed workflow begins with collecting legal documents and a simple pre-processing procedure to organize the data into an unlabeled dataset. Subsequently, the dataset is fed into a large pretrained language model on a local machine for fine-tuning. This process aims to achieve a language model tailored explicitly for generating legal document drafts, thereby enhancing the applicability of the large pretrained language model in the legal domain. The ultimate goal is to ensure information privacy and mitigate information security risks.

This paper utilizes representative open-source LLM and publicly available legal documents as experimental subjects. Detailed analysis is conducted on the text drafts generated by the model. The experimental results show that using a substantial unlabelled dataset of legal documents without Chinese word segmentation for fine-tuning a large pretrained language model on a local machine not only significantly improves the model's performance in legal document drafting tasks but also provides new insights and methods for achieving automated legal document drafting. Furthermore, this approach addresses concerns related to information privacy and enhances information security.

Our main contributions to this work can be summarized as follows:

- Using a large number of unlabeled legal documents that do not require Chinese word segmentation, fine-tuning a large-scale pre-trained language model with the smallest parameter size has achieved the goal of generating draft legal documents, reducing the computational cost of fine-tuning the language model and text generation.

- Explain how to interpret the content of the judgment from the perspective of the constituent elements specified in the legal provisions.

- This paper shows the feasibility of fine-tuning and deploying LLM on local computers and closed networks, ensuring information privacy and reducing information security risks.

- This paper is the first large-scale language model that can generate drafts of traditional Chinese legal documents for specific causes of action and also opens the training data set for subsequent researchers interested in this topic.

This paper is organized as follows: Section 2 provides the main background of LLM and text generation tasks. Section 3 explains the data sources, preprocessing, and data analysis in this paper, proposes a decomposition method based on the legal constituent elements of legal provisions, and then explains the selection of LLM and the strategy of generating text. Section 4 describes the process results of this paper's experiment. Finally, Section 5 discusses the limitations of this paper.

## 2. RELATED WORKS

In recent years, with the development of deep learning, after Vaswani published the Transformer with self-attention mechanism [5], models based on improved self-attention mechanism architecture, such as BERT, have continued to emerge and have achieved excellent performance results in many natural language processing tasks [6, 7]. The development of large-scale self-supervised learning methods is seen as a fundamental change because it allows us to train models on large unlabeled text datasets to generate a large pre-trained model that can subsequently be easily fine-tuned or prompts the model to be tuned to give good results on a variety of natural language understanding and text generation tasks [8]. Therefore, the excellent performance of the Large Language Model (LLM) in many natural language processing tasks has become a popular research direction in the field of natural language processing. GPT-3 has expanded the number of model parameters to 175 billion, and its experimental results demonstrate the robust performance of the GPT-3 large-scale language model in translation, question-answering, and word-filling tasks [3]. As mentioned above, the models have been tried to be applied to multiple tasks in the legal profession [1].

On the other hand, using LLM for legal document classification is also a research hotspot. Legal text classification is classifying legal documents into different categories, essential for legal research and practice. Existing research often uses models based on the Transformer architecture, such as RoBERTa [9], to organize legal documents. In contrast, Tagarelli et al. used the BERT model to classify legal documents and achieved preliminary research results [10].

Since legal texts are highly specialized, formalized, and diverse, applying natural language processing technology to generate legal texts is relatively complex. The traditional NLP approach also makes it challenging to deal with proper nouns, legal provisions, legal relationships, etc., in legal texts. At the same time, changes in meaning caused by subtle differences in language must also be considered. Therefore, the text generation task must overcome many difficulties when implementing text generation by applying traditional techniques. To understand these problems, some scholars have proposed using legal field corpora to fine-tune pre-trained language models [11] or legal document drafting methods based on templates and rules [12]. This method allows users to use visual tools to apply preset machine-readable rules to a preset template to generate text drafts that meet the requirements. However, these methods all face the problem of inelasticity in generating text content changes.

As mentioned above, there have been some related studies on leveraging LLMs to process legal documents, and these studies have achieved specific results. However, due to the particularity of legal texts, existing LLMs still have some limitations. First of all, existing LLMs are usually trained on a large amount of general texts [13]. Therefore, when it comes to generating particular legal texts, the text generation outcomes of these existing pre-trained models are not ideal. Secondly, legal texts contain a large number of professional terms and specific formats, and these characteristics will also affect the performance of the LLMs' text generation.

Furthermore, training or fine-tuning LLMs requires substantial computing resources. After observing the current solutions for LLMs that have been pre-trained, we will find that they often need to connect to other third-party servers to send data to other third-party servers for text generation. This text-generation mode is often unsettling for legal practitioners paying attention to information privacy and security. Therefore, there are still some challenges and difficulties in directly applying the LLM model to lawful text processing.

## 3. METHOD

### 3.1. Data Collection and Preprocessing

The first problem is the "text dataset" when leveraging natural language processing to legal documents. The Judicial Yuan has posted most of the judgments publicly online, which can be open access. After obtaining a large number of public judgments, we pre-processed the formatting of the documents one by one and extracted the full text of the judgments for subsequent training and processing.

To control the scope of text generated by the model, the judgment data used in this paper were limited to the "criminal facts (criminal description)" part of the judgment in the case of fraud. We only selected the "criminal facts" part of the judgment because this part is a stereotyped text and contains the core elements of the criminal case, including the offender's behavior, time, place, object, method, etc. Compared with other parts of the judgment, "criminal facts" usually do not involve citations and explanations of legal provisions, so they are more versatile and repetitive. Furthermore, according to the "International Classification of Crime for Statistical Purposes (ICCS)" [14], fraud, deception or corruption and other illegal acts are classified into the same type of crime. The crime There are as many as 10 types of illicit behaviors. Therefore, this paper selected the judgments of "fraud" crimes as the training dataset. In addition to the fact that the judgments of fraud cases are relatively fixed in describing criminal facts, they can provide more diverse descriptions of illegal acts. We expect to increase the diversity of content in files generated by large language models while also focusing on specific unlawful behavior.

As mentioned above, the dataset used in this paper is collected from the "fraud" case judgments published by the Judicial Yuan. To verify the feasibility of this experiment and reduce the number of calculations, we only take the "criminal facts" field of the content of the judgments. Judgment data were collected from January 1, 2011 to December 31, 2021. 74,823 verdicts as original data (judgments and rulings) were collected. To observe the relevant dataset characteristics, after removing the format and using CKIPTagger [15] for word segmentation and manual revision, there are 16,996,293 words in the training set, including 259,157 unique words. We used the TF-IDF method to analyze the nearly 60,000 judgments in the training set. The results are shown in Figure 1. We divided the original data into three parts. The training data set has 59,858 verdicts, accounting for about 80% of the original data. The remaining 20% is allocated 10% to the validation set (7,482 verdicts) and 10% to the test set. (7,483 verdicts).

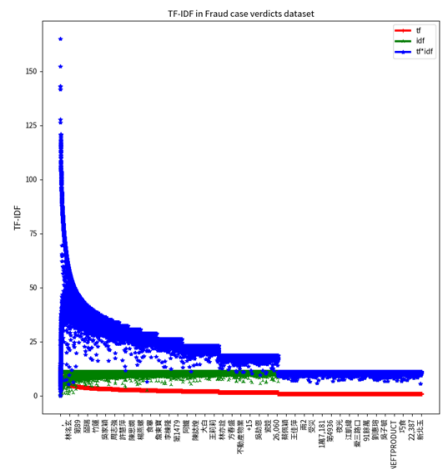

Figure 1. Apply TF-IDF to observe the data set processed by word segmentation.

## 3.2. Decomposing of the Judgments

The judgment is an essential public document made by the court. When ordinary people reading, due to a lack of legal knowledge, often read from the perspective of "somethings, people, events, sometimes, and places (what, who, when, where, how)," making it difficult to understand the connotation of the judgment. If read from the perspective of criminal law theory, it will be a theoretical process of analyzing the judge's text in the judgment and the elements of the crime stipulated in the criminal law.

Table 1 shows the provisions and translations of the Criminal Code of the Republic of China on Offenses of Fraudulence. The text without the gray mesh background is the legal effect; the punishment method after the sentence is determined. This part is outside the scope of this paper. This paper will explain the text part marked with the gray mesh, which is rich in the constitutive element of offenses of fraudulence. This paper will explain how to manually refer to the provisions of the codex article to determine whether the content format of the "criminal facts" text generated by the LLM complies with the constituent elements of the legal regulations.

Table 1. Articles the Criminal Code of the Republic of China on Offenses of Fraudulence. The provisions can be broken down into the constituent element part (with a gray background) and the legal effect part

| Offenses of Fraudulence (Chinese) | Translation |
|---|---|
| 第 339 條<br>意圖為自己或第三人不法之所有，以詐術使人將本人或第三人之物交付者，處五年以下有期徒刑、拘役或科或併科五十萬元以下罰金。 | Article 339<br>A person who by fraud causes another to deliver to him property belonging to such other or to a third person for purpose to exercise unlawful control over other's property for himself or for a fourth person shall be sentenced to imprisonment for not more than five years or short-term imprisonment; in lieu thereof, or in addition thereto, a fine of not more than five hundred thousand yuan may be imposed. |

Table 2 decomposes the part of the article Offenses of Fraudulence in Table 1: "A person who by fraud causes another to deliver to him property belonging to such other or to a third person for purpose to exercise unlawful control over other's property for himself or for a fourth person", disassembled according to the legal constituent elements. The Tag column is a customized label for this paper for the convenience of explanation.

Table 2. Decomposing of criminal law article

| The Elements of a Crime | Tag | Disassembled | Translated | Color |
|---|---|---|---|---|
| Subject of Crime | LEO_SOC | (省略) | A person (Omitted in Chinese) | ■ |
| Subjective Legal Element of the Offense | LEO_SLE | 意圖為自己或第三人不法之所有 | A person who by fraud causes for purpose to exercise unlawful control over other's property for himself or for a fourth person | ■ |
| Act | LEO_ACT | 詐術 | fraud | ■ |
| Victim/Object | LEO_VIC | 本人或第三人 | such other or to a third person | ■ |
| Causation | LEO_CAU | (陷於錯誤) | (mistaken belief) | ■ |
| Result of Hazard | LEO_ROH | 使[LEO_VIC]將[LEO_VIC]之物交付者 | causes another to deliver to him property belonging to | ■ |

Codex articles written in Chinese, the word "defendant" does not appear in the criminal code article, but the defendant's name will be clearly listed in the judgment. The victim is also presented in literal terms such as "self (本人)" and "third person (第三人)". In addition, the provisions of the offenses of fraudulence only include the result of hazard (i.e., "deliver to him property belonging to such other or to a third person (將本人或第三人之物交付)" as part of the constitutive element of causation.) to describe. Still, in practice, prosecutors or judges will add text descriptions such as "mistaken belief (陷於錯誤)" and "mistaken to be true (誤信為真)" in the criminal facts part of the indictment or judgment to match (decorate) the provisions. The word "causes another to (使人)" is used to connect the causal relationship between the defendant's fraudulent behavior and how it led to the result of hazard to the victim. The elements of the crime of fraud related to illegal conduct are more abstract in literal terms and are only described by the word "by fraud (詐術)". There are various specific illegal acts. The previous document on the "International Crime Classification for Statistics" explains this part in detail. The textual description of these illegal acts is also what we hope to get by inputting a large amount of text for LLM training and fine-tuning to learn. Then, we can generate versatility and specific text descriptions of "fraud" in various offenses of fraudulence.

犯罪事實
一、林珈羽能預見任意將所有之金融機構帳戶資料交付予他人，足供他人用為詐欺等犯罪後收受匯款，以遂其掩飾或隱匿犯罪所得財物目的之工具，詎以前開結果之發生亦不違其本意，竟基於幫助他人從事不法行為之犯意，於民國100年4月21日前之不詳時間，在不詳地點，將其向苗栗市農會所申請之帳號000000000000000號帳戶之存摺、提款卡（包括密碼），以不詳之代價，提供與不詳年籍之人使用。而該不詳年籍人士與詐騙集團成員，基於意圖為自己不法之所有，於同年月21日中午12時19分許，撥打電話向被害人張培超以假冒好友謊稱急需用錢等詐騙手法，使張培超誤以為真，而依指示操作後匯出款項新臺幣10萬元至林珈羽上開帳戶內而受騙。
二、案經張培超訴由苗栗縣警察局苗栗分局報告偵辦。

Figure 2. Read the criminal facts from the perspective of legal constitutive elements.

Figure 2 shows the criminal facts in a certain fraud judgment. It is a codex article analyzing the criminal facts from the perspective of the constituent elements of criminal law. It can be disassembled using the method of Table 2 mentioned above. After decomposing, it is shown in Table 3. The left column of Table 3 is an abstract description of the legal constitutive elements of the crime. You can refer to the explanation in Table 2 in the previous paragraph. The "Offenses of Fraudulence" legal provisions of Criminal Law are based on relatively specific descriptions of it with text, such as the contents of the disassembled column in Table 2. Here, we decompose the criminal facts in Figure 2 and map them to each component, as shown in Table 3. In Figure 2, you should find that some words are not marked. Those words are generally connectives and narrative sentences and have nothing to do with the legal constitutive elements. For example, 「竟」、「詎」、「而」, etc.

Table 3. Legal constitutive elements and actual examples.

| The Legal Constitutive Elements | Tag | Example | Color |
|---|---|---|---|
| Subject of Crime | LEO_SOC | 林珈羽、該不詳年籍人士、詐騙集團成員 | ■ |
| Subjective Legal Element of the Offense | LEO_SLE | 預見[LEO_ACT]、意圖為自己不法所有 | ■ |
| Act | LEO_ACT | 撥打電話向[LEO_VIC]以假冒好友謊稱急需用錢等詐騙手法 | ■ |
| Victim/Object | LEO_VIC | 被害人張培超 | ■ |
| Causation | LEO_CAU | 使[LEO_VIC]誤以為真 | ■ |
| Result of Hazard | LEO_ROH | 依指示操作後匯出款項新臺幣 10 萬元至[LEO_VIC]上開帳戶內 | ■ |

Observing Figure 2 and decomposing the judgment in the manner explained above as an example, we can get the basic order of the literal appearance of each legal constitutive element when writing a "criminal fact" of the offenses of fraudulence as follows:

1. Subject of Crime (LEO_SOC)
2. Subjective Legal Element of the Offense (LEO_SLE)
3. Act (LEO_ACT)
4. Victim (LEO_VIC)
5. Causation（LEO_CAU）
6. Result of Hazard（LEO_ROH）

The order where the relevant legal constituent elements appear in the above draft format is represented by symbols as follows:

$$LEO\_SOC \rightarrow LEO\_SLE \rightarrow LEO\_ACT \rightarrow LEO\_VIC \rightarrow LEO\_CAU \rightarrow LEO\_ROH$$

Equation 1

We replace the example in Figure 2 with the symbols of Equation 1, and we can obtain the results shown in Table 4.

Table 4. The contents of Figure 2 are represented by Equation 1.

> 一、<LEO_SOC>能<LEO_SLE><LEO_ACT>，詎<LEO_SLE>，竟<LEO_SLE>，於民國 100 年 4 月 21 日前之不詳時間，在不詳地點，<LEO_ACT><LEO_SOC><LEO_ACT>。而<LEO_SOC>與<LEO_SOC>，基於<LEO_SLE>，於同年月 21 日中午 12 時 19 分許，<LEO_ACT><LEO_VIC><LEO_ACT>， <LEO_CAU><LEO_VIC><LEO_CAU>， 而 <LEO_ROH><LEO_SOC><LEO_ROH>而受騙。

Observing the literals presented in Table 4, the content of the generated text includes literals such as time, place, and person's name. Because they are randomly and automatically generated by LLM, in actual use, they must be modified by the user according to the facts. Excluding the above limitations, we can assume that as long as the draft document generated by the LLM has the literals of each legal constitutive element in the order of Equation 1, we can be sure that the draft format generated by LLM is correct. Suppose a more relaxed evaluation standard is adopted, as long as those above six legal constituent elements appear literally in the generated text. In that case, the draft format can be determined to be correct. In the following content, we will evaluate the format correctness of the generated draft legal text and use the disassembly mentioned above method to assess it.

### 3.3. LLM Selection and Text-Generation Strategies

The basic model of the experiment in this paper is the pre-trained large-scale language model BLOOM [16] using the Casual Language Model. The consideration for choosing BLOOM is not only because it is an open-source LLM but also because more than 46 language texts, including traditional Chinese text, are used when training it. This multi-language training corpus makes BLOOM versatile. Because high-resource languages help improve the performance of low-resource and very-low-resource languages, they can be directly applied to Traditional Chinese tasks [17]. BLOOM models are available in various parameter sizes, ranging from 350M to 176B. In the early stages of the experiment, we first entered prompts directly on the 560M parameters model. We tried to let BLOOM generate a draft of criminal facts about offenses of fraudulence. The result was quite unsatisfactory, but this was also the expected result.

We loaded the BLOOM model with 560M parameters on the local computer for fine-tuning, which computer equips RTX-3090, and the dataset was also preprocessed on the local computer, as described in this section. Because this processing and training process does not send data to external processing, and the training process is also performed locally, this process allows users sensitive to generating legal text drafts to be protected in terms of information privacy and reduces the risks of Information security. The process of fine-tuning the model is briefly described in Figure 3.

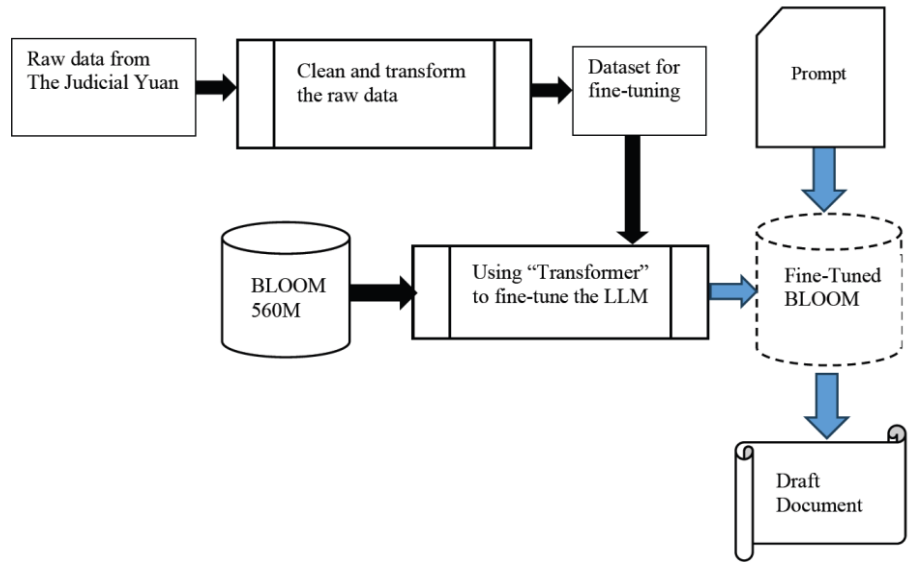

Figure 3. The process of fine-tuning LLM and generating drafts.

The most commonly used metric for evaluating the performance of text generation is perplexity. The experiments of this paper found that the perplexity is not very helpful in evaluating the performance of generating legal draft documents. On the other hand, the number of rounds (epoch) during fine-tuning does not require too many rounds. Too many rounds will increase confusion and increase the time cost of training. This part will be shown in the experimental results section.

At present, there are various strategies for generating text, such as Greedy Search, Beam search [18], Top-K sampling, Top-p (nucleus) sampling [19], etc., which are common. This paper uses the above strategies to generate text and review the results. Finally, the strategy of Top-K sampling combined with Top-p sampling is selected because the text generated in this way is the most grammatically expected.

### 3.4. Software Tools Selection

The programming language used in this paper is Python, mainly using the "transformers" package to fine-tune BLOOM. Because the verdicts have their own specified format, we wrote our pre-processing program to build the dataset needed for our experiment, combined the data and stored it in Hugging Face. When fine-tuning the model, use the "datasets" package to process the dataset placed on Hugging Face. When fine-tuning BLOOM, the GPU we used was an nVidia RTX 3090, so we also needed to use the CUDA toolkit.

## 4. EXPERIMENTS RESULT

Before fine-tuning BLOOM, we first directly give a sentence extracted from the judgment as a prompt (the name is fictitious): "Que Heng Da clearly knows that the passbook, ATM card and password of the financial account are essential financial management tools for his own use." (關很大明知金融帳戶之存摺、提款卡及密碼係供自己使用之重要理財工具) is used as the starting prompt to test the text generation results of the model without fine-tuning. Table 5 is the result generated using the three strategies mentioned in this paper and inputting the above prompt to BLOOM with 560M parameters.

Table 5. Results after inputting prompts on BLOOM without fine-tuning.

| Text-Generation Strategy | Text Generation |
|---|---|
| Greedy Search | 闕很大明知金融帳戶之存摺、提款卡及密碼係供自己使用之重要理財工具，故於本網站之「個人資料保護」部分，特別說明個人資料之收集、使用、保管、保護及保密等相關規定，並說明本網站之「個人資料保護」部分之使用方式及相關規定。</s> |
| Beam Search | 闕很大明知金融帳戶之存摺、提款卡及密碼係供自己使用之重要理財工具，故於本網站之「個人資料保護政策」下，特說明以下事項：<br>1. 本網站所蒐集之個人資訊，僅供本公司之內部使用，並僅能用於本公司的內部管理，不得用於任何第三者之用途。若您有其他任何疑問，請與我們聯繫，我們將盡力協助您解決問題。</s> |
| Sampling Top-k + Top-p | 闕很大明知金融帳戶之存摺、提款卡及密碼係供自己使用之重要理財工具，因此亦已開始實施新帳戶管理制度。此制度於 2013 年 5 月 1 日正式實施，該制度亦成為美國首次納入的金融دشتن法。根據美國銀行業的統計顯示，美國共有超過 30 億戶使用者，其中 20%至 40%皆通過使用銀行帳戶進行交易，而銀行帳戶的存摺、提款卡、密碼亦經常被使用。因此美國銀行業也開始進行新帳戶的管理，以便為使用者節省時間和費用，提高金融帳戶的使用效率。「銀行帳戶的存摺、提款卡及密碼」這三個帳戶，是美國銀行業管理制度中最為重要的帳戶，同時也是目前使用銀行帳戶進行交易的重要保障，對於美國銀行業的交易安全，重要性之所在不言而喻。由於銀行帳戶是美國銀行業最重要的帳戶之一，為了確保使用者使用銀行帳戶進行交易的安全，美國銀行業將「銀行帳戶的存摺、提款卡及密碼」這三個帳戶，當使用者使用完畢後，銀行業會將其移除並另換其他的帳戶。藉由銀行帳戶的存摺、提款卡及密碼這三個帳戶進行交易，銀行業可以節省不少的時間與費用，例如由於銀行帳戶的存摺、提款卡及密碼被其他使用者使用，銀行業會收取費用，若使用者無法使用銀行帳戶進行交易，銀行業也會收取費用。 因此由於銀行帳戶的存摺、提款卡及密碼是美國銀行業管理制度中最重要的帳戶之一，銀行業為了保障使用者使用銀行帳戶進行交易的安全，銀行業特別制定了「銀行帳戶的存摺、提款卡及密碼」新帳戶管理制度。銀行業針對此制度特別制定了管理制度 |

From the results in Table 5, the original model without fine-tuning, no matter which text generation strategy is used, cannot generate the draft content of the "criminal facts" of the judgment, as shown in Figure 2.

At the beginning of the experiment in this paper, a tiny model with a parameter scale of 560M was used for fine-tuning. In the environment of nVidia RTX 3090 with 24GB RAM, it took about four hours to complete 5 epochs of training using the "criminal facts" of 59,858 verdicts dataset.

When fine-tuning a small model with 1.1 billion parameters, the difficulty encountered is insufficient computing power. Initially, we tried fine-tuning in an environment equipped with nVidia RTX 3090 and 24GB RAM. Still, no matter how we adjusted the dataset's size or the various hyper-parameters of the training language model, we could not successfully deploy the small model with 1.1 billion parameters in the computing environment above. To speed up the fine-tuning process, we changed to the A100 environment, which was equipped with 40GB RAM to fine-tune the model, and we finally obtained a small model with a perplexity of 4.71.

After fine-tuning, we released the model to Hugging Face. The demonstration platform shows two application approaches of this LLM in writing draft legal documents: writing assistance and entire draft text generation. Figure 4 is a sample text generated from a random complete draft.

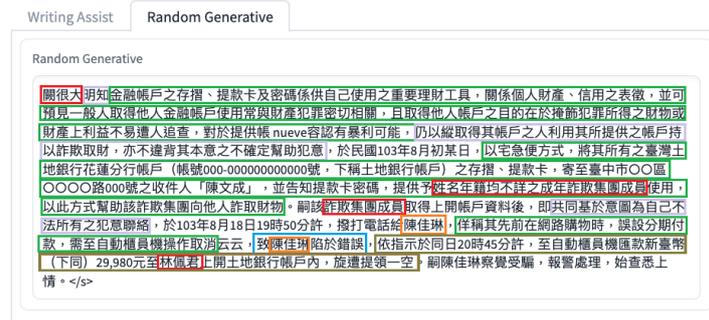

Figure 4. An example of the fine-tuned LLM generates a text draft after entering a prompt.

Figure 4 is the prompt we entered into the LLM after fine-tuning this paper: "Que Heng Da clearly knows that the passbook, ATM card and password of the financial account are important financial tools for his own use," （闕很大明知金融帳戶之存摺、提款卡及密碼係供自己使用之重要理財工具，）, using Top-K and Top-p sampling strategies, and limit the upper limit of generated tokens to 500.

We use the decomposing method described in the previous section and apply Equation 1 to decompose the results in Figure 4. We can obtain the decomposed results in Table 6.

Table 6. Apply Equation 1 to decompose the results in Figure 3.

<LEO_SOC><LEO_SLE><LEO_ACT>，<LEO_SLE>，於民國 103 年 8 月初某日，<LEO_ACT><LEO_SOC><LEO_ACT>。嗣該<LEO_SOC>取得上開帳戶資料後，即<LEO_SLE>，於 103 年 8 月 18 日 19 時 50 分許，撥打電話給<LEO_VIC>，<LEO_ACT> 云 云 ， <LEO_CAU><LEO_VIC><LEO_CAU>，<LEO_ROH><LEO_SOC><LEO_ROH>。嗣<LEO_VIC>察覺受騙，報警處理，始查悉上情。

Checking the order of appearance of each label in Table 6, it is indeed the order shown in Equation 1, that is, "Defendant's name (subject of crime) (LEO_SOC) -> Subjective Legal Element of the Offense (LEO_SLE) -> Illegal act (LEO_ACT) - >Victim (LEO_VIC) -> Causation (LEO_CAU) -> Result of Hazard (LEO_ROH)" to generate a draft of "criminal fact".

Table 7 shows the loss and confusion when training the 560M parameters model. We can find that increasing the number of training rounds will not reduce the perplexity of the model in text generation but will tend to increase significantly.

Table 7. Fine-tune epoch and perplexity

| Epoch | Training Loss | Validation Loss |
|---|---|---|
| 1 | 1.657000 | 1.711774 |
| 2 | 1.447700 | 1.646734 |
| 3 | 1.235100 | 1.673958 |
| 4 | 0.990400 | 1.829225 |
| 5 | 0.766500 | 2.141219 |
| 6 | 0.669200 | 2.338404 |
| 7 | 0.486300 | 2.767606 |
| 8 | 0.338900 | 3.110578 |
| 9 | 0.217200 | 3.391885 |
| 10 | 0.141600 | 3.577524 |

Epoch=5, Perplexity: 8.51 Epoch=10, Perplexity: 35.78

## 5. DISCUSSION AND FUTURE WORKS

Based on the results of this paper, we can draw the following conclusion: It is feasible to fine-tune and deploy LLMs on the local computer to generate legal document drafts and perform excellently in experiments. In addition, we also found that as long as a sufficient amount of legal professional texts are collected to fine-tune an LLM, the performance of the text generated by the model can be further improved. These results show that large language models have extensive application prospects in generating legal document drafts.

When fine-tuning a large language model, the adequacy of hardware resources is an important consideration. Especially when fine-tuning larger models, higher GPU computing power and GPU memory space are required. Relevant research also points out that the GPU's computing power and memory size significantly impact fine-tuning [20]. When fine-tuning large models, if more GPUs can be connected in series, the training process can be accelerated considerably and improve the model's performance. This paper shows that using a large number of unlabeled legal documents that do not require Chinese word segmentation and using consumer-grade GPUs to fine-tune a large-scale pre-trained language model with the smallest parameter size has achieved the goal of generating draft legal documents, reducing the computational cost of fine-tuning the language model and text generation.

It is worth noting that our study has several limitations. First, our experimental dataset is still relatively small, so more data are needed to validate the robustness of our conclusions. Secondly, we use a single perplexity as the evaluation metric, which may not fully reflect the model's performance in actual applications. Future research can further explore other evaluation metrics to evaluate model performance comprehensively. Overall, our study provides valuable experience and inspiration for applying LLMs in legal document draft generation tasks and provides a reference for further exploring the application of LLMs in other application scenarios. In future research, these results will play a positive role in promoting the development of related fields.

The problem that this paper does not solve is the "hallucinations problem" of natural language text generation that occurs in large language models [21]. This can be seen from the experimental results in Figure 3 that the model generated the word: "nueve", which is irrelevant and does not exist in the training data set. This part can be left to research further to solve this problem.

The model fine-tuned in this paper is publicly available for download and use at https://huggingface.co/jslin09/bloom-560m-finetuned-fraud. Its online demonstration can be found at https://huggingface.co/spaces/jslin09/legal_document_drafting operate. The dataset used to fine-tune the model in this experiment is public at https://huggingface.co/datasets/jslin09/Fraud_Case_Verdicts.

pp. 1-38.

**AUTHORS**

**Chun-Hsien Lin** is currently a Prosecutorial Affairs Officer at the Taiwan High Prosecutors Office and pursuing a Ph.D. degree from the National Taiwan University of the Department of Computer Science & Information Engineering. His research is mainly on AI and law, natural language processing, machine learning, etc.

**Pu-Jen Cheng** received his Ph.D. in Computer Science at National Chiao Tung University in 2001. He went to the Institute of Information Science, Academia Sinica, as a Postdoctoral Fellow for more than four years. Starting from August 2006, he joined the Department of Computer Science and Information Engineering faculty at the National Taiwan University and was also jointly appointed at the Graduate Institute of Networking and Multimedia, National Taiwan University. He is a member of the ROC Phi Tau Phi Scholastic Honor Society and the ACM/SIGIR

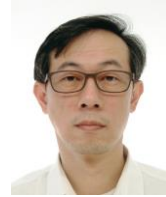

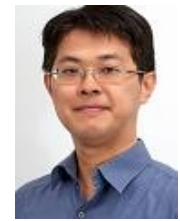